\documentclass{article} 
\usepackage{bm}
\usepackage{arxiv}
\usepackage[switch]{lineno}
\usepackage{booktabs}
\usepackage{multirow}
\usepackage{amsfonts,amssymb}
\usepackage{amsmath}
\usepackage{graphicx}
\usepackage{subcaption}
\usepackage[colorlinks,
           linkcolor=blue,
           anchorcolor=blue,
           citecolor=green
           ]{hyperref}
\newtheorem{myDef}{Definition}

\usepackage{url}

\title{A Deep Graph Neural Networks Architecture Design: From Global Pyramid-like Shrinkage Skeleton to Local Topology Link Rewiring}

\author{
  Gege~Zhang \\
 Department of Automation\\
 Shanghai Jiao Tong University \\
Shanghai 200240, PR China\\
  \texttt{ggzhang@sjtu.edu.cn} \\
  }

\begin{document}
\maketitle
\begin{abstract}
   Expressivity plays a fundamental role in evaluating deep neural networks, and it is closely related to understanding the limit of performance improvement.
   In this paper, we propose a three-pipeline training framework based on critical expressivity, including global model contraction, weight evolution, and link's weight rewiring.
   Specifically, we propose a pyramidal-like skeleton to overcome the saddle points that affect information transfer.
   Then we analyze the reason for the modularity (clustering) phenomenon in network topology and use it to rewire potential erroneous weighted links.
   We conduct numerical experiments on node classification and the results confirm that the proposed training framework leads to a significantly improved performance in terms of fast convergence and robustness to potential erroneous weighted links.
   The architecture design on GNNs, in turn, verifies the expressivity of GNNs from dynamics and topological space aspects and provides useful guidelines in designing more efficient neural networks.
\end{abstract}

\section{Introduction}
Deep neural networks (DNNs) have achieved an outstanding performance for various learning tasks such as speech recognition, image classification, and visual object recognition etc.\cite{lecun2015deep}.
It is well known that DNNs can apprioximate almost any nonlinear functions and make end-to-end learning possible \cite{mallat2016understanding, RaghuPKGS17}.
Most recently, there has been a surge of interest in graph neural networks (GNNs) since they can capture the dependency of graphs by accounting for the message passing between nodes \cite{zhou2018graph, abs-1901-00596}.
This appealing feature has renewed interest in answering a variety of fundamental questions involving the interpretation, generalization, model selection, and convergence of GNNs (DNNs) \cite{NovakBAPS18}.

When developing innovative GNN techniques, it is imperative to explore the physical and mathematical principles that explain the observed phenomenon, which ultimately provides guidelines for creative designs.
There is a rich literature studying the effectiveness of GNNs from various aspects \cite{RaghuPKGS17, DongSZZ17}. \footnote{This field is called the interpretability or expressivity, and here we use the later terminology to encompass all efforts in this area.}
For example, Scarselli et al. \cite{Scarselli2009Computational} first showed that GNNs can approximate a large class of functions in probability.
Kawamoto et al. \cite{KawamotoTO18} provided a theoretical analysis of GNNs based on mean-field theory for graph partitioning tasks.
Lei et al. \cite{lei2017deriving} designed a recurrent neural architecture inspired by graph kernels and discussed its equivalence between Weisfeiler-Lehman kernels.
Moreover, Xu et al. \cite{xu2018powerful} proved that the expressivity of GNNs was as powerful as that of the Weisfeler-Lehman graph isomorphism test.
These efforts have deepened our understanding of the expressive power of GNNs, however, general guidelines are still largely needed for designing better neural architectures and overcoming issues in the training of neural networks, such as parametric choices (e.g., width and depth), vanishing, and exploding gradient problems.

Based on the previous research regarding the expressivity of DNNs \cite{RaghuPKGS17, zhang2019expressivity}, we understand that DNNs use the spatial space that offers an informational representation and evolve toward a critical state (i.e., critical points $\bm{0}$) as the depth increase, corresponding to increasing informational entropy. The main goal of this paper is to further study the optimal topology design on GNNs based on the criticality theorem. The main contributions of this paper can be summarized as follows:
\begin{itemize}
  \item We propose a tree-pipeline training framework involving global skeleton determination (width determination) and local topological link rewiring.
  \item The weak improvement of current pruning algorithms is examined, in which a rank-constrained or sparsity-constrained regularization imposed on non-convex optimization will prevent the tendency toward the critical state $\bm{0}$.
  \item The modularity (clustering) phenomenon in network topology is utilized for erroneous weight rewiring in the weight matrix, which in turn verify the modularity phenomenon in GNNs.
\end{itemize}

\section{Graph Neural Network and Its Critical Representation}\label{sec:GraphNeuralNetworks}
The goal of this paper is to explore the representation capabilities of GNNs on graphs.
To this end, we consider a vanilla GNN with feedforward dynamics.
Suppose the input graphs are characterized by a vertex  set of size $V$ and a $D$-dimensional feature vector with elements $X_{iu} (i\in V, u\in \{1,\dots, D\})$,
then the state matrix $\bm{X} =[X_{iu}]$ is given by
 \begin{equation}\label{eq:GNNModel}
X_{i u}^{t+1}=\sum_{j v} \phi\left(A_{i j} X_{j v}^{t}W_{v u}^{t}\right) +b_{i u}^{t},
\end{equation}
where $\phi(\cdot)$ is a non-linear activation function,  $\bm{A}=[A_{ij}]$ is the adjacency matrix of network topology, $\bm{W}^{t}=\left[W_{vu}^{t}\right]$ is a linear transformation of feature space, $\bm{b}^{t}=\left[b_{i u}^{t}\right]$ is a bias term, and the layer is indexed by $t \in \{1,\dots, T\} $.

The general idea behind GNNs is that nodes can be recursively aggregated and propagated to the next layer for complex calculations.
In this respect, the network structure of neural networks is typically described as graphs in which nodes act as neurons, and each edge links the output of one neuron to the input of another. Graph matching refers to a computational problem of establishing a one-to-one bijective correspondence between the vertex set of graphs.
Therefore, graph matching between a pair of graphs is analogous to representing graphs using GNN \cite{LiGDVK19}.
In the next, we discuss the dynamics aspects of the graph matching.

Based on the Banach Fixed Point Theorem in dynamics, we know that the unique solution of differential equations in (\ref{eq:GNNModel}) can be obtained through an iterative process
\begin{equation}\label{eq:RecursionDynamics}
{{\bf{X}}^{t + 1}} = \phi(\bm{A}\dots\phi(\bm{A}\phi\left( \bm{A}{{{\bm{X}}^1}{\bm{W}^1}} \right){\bm{W}^2})\dots{\bm{W}^t}).
\end{equation}
To prevent the system from being chaotic, the eigenvalue of hidden states should satisfy  $|\lambda_i(\bm{X}^t)|)<1, i\in {1,\dots,\infty}$.
Assume that the weight matrix $\bm{W}^t$ is randomly distributed. Then both forward propagation and backpropagation are the information transfer powered by dynamics from $[\lambda_1(\bm{X}^t),\dots,\lambda_i(\bm{X}^t),\dots]$ toward the critical state, i,e., critical points $\bm{0}$, which has abundant expressivity \cite{zhang2019expressivity}.
The results can be generalized to local topological vector spaces via Schauder fixed point theorem \cite{Bonsall1962}.
The theorem illustrates that there always exists a fixed point if $X$ is a closed convex subset of local topological space $S$ and $f$ is a continuous self-mapping such that $f(X)$ is contained in a compact subset of $X$.
In this respect, training a GNN is to construct an inexact graph matching through convex-relaxation.
In the next sections, we will further demonstrate this point in more detail.
\section{The Training Issues and Global Skeleton in Graph Neural Networks}\label{sec:SkeletonDesign}

From the Schauder fixed point theorem, to reach the critical state, one should construct a convex network structure and an input convex topological space.
First, we examine the topological structure issue.
The current training methods in GNNs are mainly based on backpropagation, including those gradient-based methods.
However, ordinary gradient descent cannot guarantee convergence to the global minimum, since the cost function is always non-convex.
Another impediment to the convex optimization is the presence of saddle points in high dimensional representation.
The current network structure is pre-set before training, and usually over-parameterized, which may generate many saddle points \cite{ChoromanskaHMAL15}.
In addition, (\ref{eq:RecursionDynamics}) suggests that a global minimum in low dimension may attenuate to a saddle point $0$ in a high dimensional setting by layer-wise multiplying $\lambda_i$, the so-called proliferation of saddle points \cite{DauphinPGCGB14}.

Mathematically, to determine whether a solution is a local minimum, a global minimum or a saddle point, one needs to calculate the eigenvalues of its Hessian matrix at any given point. If all the eigenvalues have both positive and negative values, there will also be a zero value, corresponding to a saddle point.
If all the eigenvalues are positive at any point, there exists a global minimum. Although some recent work addressed this issue either by adopting noisy stochastic gradient descent (SGD) or second-order Hessian information (e.g., Adam), they only avoided the local minimums, and the saddle point issue remains unresolved.

In addition to the backpropagation, another popular method for solving non-convex optimization is the alternating direction method (e.g., PARAFAC for matrix/tensor decomposition) \cite{cichocki2016tensor, AghasiANR17}, which is an alternating matrix optimization algorithm that solves optimization problems by breaking down the convex optimization into smaller parts. Taylor et al. \cite{taylor2016training} pointed out that the alternating direction method of multipliers (ADMM) could overcome the gradient vanishing or explosion issue in backpropagation, and could be implemented in parallel and distributed computing environment.
However, the theoretical understanding of the convergence of ADMM remains challenging when the objective function is non-convex, and simulation examples showed that ADMM could achieve high precision very slowly \cite{BoydPCPE11}.

To avoid the saddle points caused by over-parameterization and high-dimensional representation, it is recommended that the network structure should have a pyramid-like shrinkage property \footnote{Some literature calls it as model compression, here we are prone to dimensional contraction to describe the relationship between successive layers.}.
The shrinkage characteristics refer to the situation that the width of the next layer needs to shrink down compared to the current layer.
Specifically, a network structure is typically characterized by the width (i.e., the number of nodes in each layer) and the depth  (measured by the number of hidden layers)  of GNNs. In theory, the depth relies on the time dependence and period of data itself. Hence there is no definitive way to determine the optimal value for depth given a specific dataset, and this is usually obtained by numerical trials. Therefore we here mainly focus on estimating the width.
In mathematics, given a complete input, network width can be determined by identifying the latent rank of the observable matrix. This field is called low-rank recovery (or low-rank matrix completion).
By imposing a rank constraint at each layer, the network width should show a pyramid-like structure.
We provide more details about low-rank recovery in Appendix, and examine the proposed hypothesis via simulation experiments.
\subsection{What is Wrong with Existing Pruning Algorithms}\label{sec:PruningAlgorithm}
This section discuss the criticality issue by examining the current pruning algorithms.
Initially, Denil et al. showed in Denil et al. \cite{DenilSDRF13} that there was a considerable redundant structure in existing networks.
To reduce the number of parameters and nodes, researchers have developed various network pruning algorithms to eliminate unnecessary connections or neurons without negatively affecting convergence. A typical pruning algorithm has a three-stage pipeline, i.e., 1) training a large, over-parameterized model;
 2) pruning the trained over-parameterized model according to specific criteria; 3) fine-tuning the pruned model to regain the optimal accuracy.
The core pruning procedure is divided into three categories: weight pruning, structured pruning, and layer pruning.
Since the layer pruning depends on the matching between the model and actual data, this paper focuses on the first two pruning techniques.
Weight pruning also learns networks by adding sparsity or rank constraints on GNNs, i.e.,
\begin{equation}
     \bm{W}=\mathop{\arg\min}_{\bm{W}} (\bm{X}^{t+1}- \phi(\bm{A}\bm{X}\bm{W}^t)+\lVert\bm{W}^t\lVert).
\end{equation}
From the critical analysis, this constraint by imposing regularization will reverse the tendency toward criticality when the network approaches the critical state  $\bm{0}$. From a searching perspective, by mixing up the topology search with weight evolution in one model, the resulting algorithm cannot achieve representation with high precision. Liu et al. \cite{LiuSZHD19} also showed in an experimental analysis that current pruning algorithms only gave a comparable or worse performance than training models with randomly initialized weights. They also emphasized that the pruned architecture, rather than ``significant'' weights, was more important in improving convergence, which is consistent with our analysis.
\section{Robust Topological link Rewiring}
As mentioned earlier, the assumption in global network skeleton design is built on a complete observation of the input  $\bm{X}$. In real-world scenes, however, graphs often suffer from the missing edge or missing node features, and the inputs are incomplete \cite{DavenportR16}.
Besides, specific-task based backpropagation learn quickly from current inputs and may ``forget" the previous learning experience.
 As a result, the potential accumulated erroneous inputs may eventually form an erroneous topology structure \cite{nelwamondo2007missing}.
In such settings, we need to recover a complete and accurate network topology via a robust design.
Therefore, this section introduces a robust topological design for potential erroneous wights.

A classical approach for increasing network robustness is the use of local (geometric) topological structures. An intuitive understanding of the topological robustness is to provide path redundancy between vertices.
When one path fails, communication can continue through other alternative routes.
Besides, the experiments visualizing the hidden states during training also observed a growing modularity or clustering phenomenon  \cite{KawamotoTO18,hou2020learning}.
This phenomenon generally appears in the real-world coupled systems consisting of dynamics and local topological structures \cite{li2010global}.
In all, this general phenomenon implies one can rewire the possible erroneous links by exploiting the local topological structures as an informational redundancy for self-checking.

The modularity presented in the network topology can be viewed as a cluster consensus that each cluster consists of multiple interacting intelligent agents,
and training the network topology is a process of building consensus among each cluster.
Most consensus problem would converge to the average (proof is given in the Appendix), that is, the current state of each agent is an average of local objective function
\begin{equation}
   \min \frac{1}{n} \sum_{i=1}^{n} f_{i}(x_i)=\sum_{j=1}^n A_{ij}x_j(t), i=1,\dots,n, \quad \bm{x_i} \in \mathcal{X},\\
\end{equation}
where $f_i(\cdot)$ is the loss function corresponding to agent $i$, and $x \in \mathcal{X}$ is an unknown state to be optimized.
Since network topology in GNNs can be viewed as a graph, its convergence can be handled via graph theory.

For weights in GNNs, there are both positive and negative values. For an undirected graph with all positive weights, that belongs to a class of $Z$ matrix admitting many favorable properties, has been widely studied.
For example, the spectrum of the positive weighted graph Laplacian $\mathfrak{S}(\bm{L})$  has the form: $\mathfrak{S}(L)=\left\{0=\lambda_{1} \leq \lambda_{2} \leq \cdots \leq \lambda_{\infty}\right\}$.
The second smallest Laplacian eigenvalue $\lambda_2(\bm{L})$ is considered as a measure of algebraic connectivity on graphs.
For directed graphs, algebraic connectivity also holds (proof is given in the Appendix).
The consensus can be reached when all weights within a connected graph are positive.
In contrast, negative weights indicate an antagonistic or anticorrelated interaction between nodes.
The existence of both positive and negative weights in the neural architecture may lead to network modularity (clustering) \cite{zelazo2014definiteness}.
The consensus of a graph with negative weights relies on the specific algebraic connectivity measure.
On the other hand, one can make graph cuts or graph partitioning in which the link with positively weighted edges is within one module and the negative ones are between modules.

Given the modularity feature exhibited in the evolutionary dynamics, an intuitive idea is to exploit local connectivity as redundant information to fine-tune the local link during training.
Since the original weights are in general randomly generated, and the algebraic connectivity increases monotonously to form clustering,
one can impose the algebraic connectivity based regularization on the loss function after several epochs waiting for the cluster forming \cite{tam2020fiedler}
\begin{equation}\label{eq:coupledConstraint}
\min _{\bm{W}} \mathcal{L}\left(Y, \hat{\bm{f}}_{\bm{W}}(X)\right)+\delta \lambda_{2}(|\bm{L}|),
\end{equation}
where $\bm{L}$ is the Laplacian matrix converted from the weight matrix $\bm{W}$, $\lambda_{2}(|\cdot|)$ is the Fielder value of the graph of each cluster, and $\delta$ is a tuning parameter.

By imposing the regularization term in the loss function, (\ref{eq:coupledConstraint}) becomes less transparent to observe the specific erroneous links.
Meanwhile, the link should be pruned to exert a localized influence, i.e., the regularization imposed on the overall topology may offset the effects of local link rewiring.
To achieve a better interpretation of the results, we choose to use a greedy algorithm to verify our hypothesis,
rewire possible erroneous links and better understand the clustering phenomena in the training procedure.
We discuss the localized link rewiring in GNNs in the next section.

\subsection{Link's Weight Rewiring to Enhance Algebraic Connectivity}\label{subsec:Rewiringedges}
\begin{figure*}[htpb!]
    \centering
    \includegraphics[width=\textwidth]{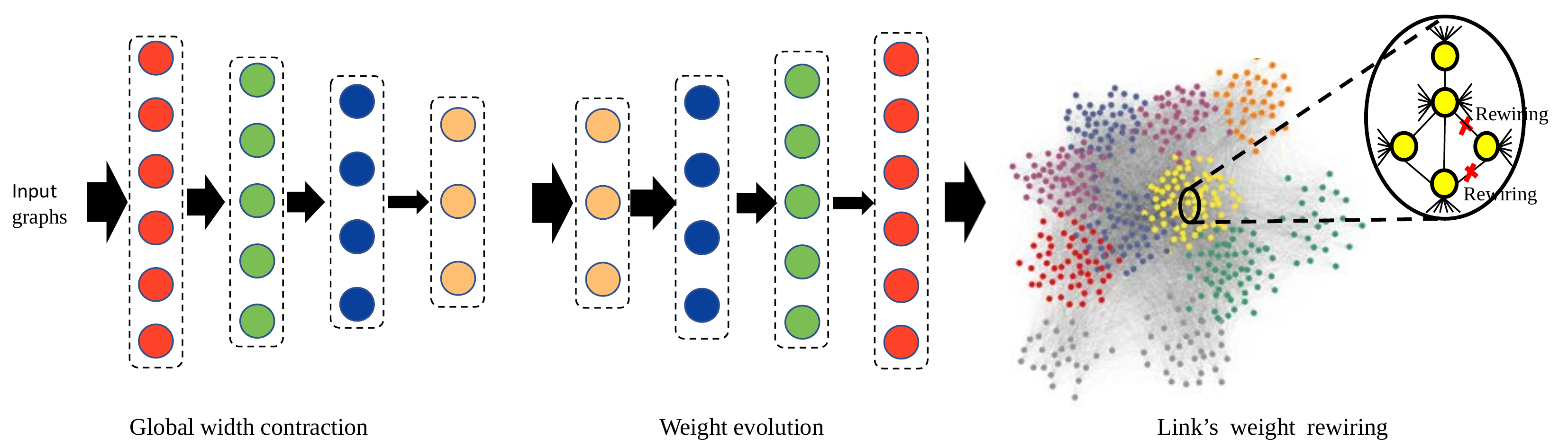}
    \caption{A graph neural network architecture design.
    The tree-pipeline successively includes global width contraction, weight evolution, and link's weight rewiring.
    In the initial dynamics, information transfer is from front to back.
In supervised learning, information transfer is in the opposite direction since it is subject to specific task constraints imposed by outputs,
accompanied by declining transfer capacity in backpropagation caused by numerous local minimums or saddle points.
To accelerate the informational transfer, the global architecture should have a pyramid-like shrinkage shape to prevent the saddle points caused by over-parameterized settings.
After the weight evolution forming the modularity in topological structure, one can use the topological structure as redundant information to rewire possible erroneous topological links.
}
    \label{fig:Frameworks}
  \end{figure*}
For a disconnected graph, its algebraic connectivity is $0$, and one can increase the algebraic connectivity by rewiring links.
Note that in addition to the adjacency matrix, incidence matrices can also be used to reprensent a gragh
\begin{equation}\label{eq:LaplacianIncidenceMatrix}
\mathbf{L}=\mathbf{H} \mathbf{H}^{T}=\sum_{l=1}^{m} \mathbf{h}_{l} \mathbf{h}_{l}^{T}.
\end{equation}
$\bm{H} = [\bm{h}_1, \dots, \bm{h}_m]\in \mathbb{R}^{n\times m}$ is the node-edge incidence matrix of graph $\mathcal{G}_{\text{sub}}$, and each edge vector $\bm{h}_l$ denotes vertex $V_i$ joining with vertex $V_j$ whose entries are $[h_l]_i = 1, [h_l]_j =-1$ and $0$ elsewhere.
Given an initial  graph $\mathcal{G}_0$, the connectivity of weighted Laplacian matrix $\bm{L}_0$ can be increased by adding new edges
\begin{equation}\label{eq:addingEdges}
\bm{L}(x)=\bm{L}_{0}+\sum_{l=1}^{L} \beta_l w_{l} \bm{h}_{l} \bm{h}_{l}^{T},
\end{equation}
where $\beta_l \in \{0, 1\}$ is a boolean variable indicating whether the $l$th edge is selected, and $w_l$ is the weight being added to edge $l$.
If edge $l$ is added to graph $\mathbb{G}$,
the partial derivative of $\lambda_2(\bm{L}(\beta))$ with respect to $\beta_l$ gives the first order approximation of the increase of $\lambda_2(\bm{L}(\beta))$.
 According to the algebraic connectivity of directed graphs in Appendix, we have
 \begin{equation}\label{eq:derivative}
 \frac{\partial}{\partial \beta_l \lambda_{2}(L(\beta))}=\bm{v}^{T} \frac{\partial L(\beta)}{\partial \beta_{l}} \bm{v}.
 \end{equation}
 Substitute  (\ref{eq:addingEdges}) into (\ref{eq:derivative}), we obtain
  \begin{equation}\label{eq:ChoosingEdges}
  \begin{aligned}
     & \frac{\partial}{\partial \beta_{l}} \lambda_{2}(\bm{L}(\beta)) \\
     =& \bm{v}^{T} \frac{\partial\left(\bm{L}_{0}+\sum_{l=1}^L\beta_{l} w_{l} \bm{h}_{l} \bm{h}_{l}^{T}\right)}{\partial \beta_{l}} \bm{v} \\
     =& \bm{v}^{T}\left(w_{l} \bm{h}_{l} \bm{h}_{l}^{T}\right) \bm{v}=w_{l}\left(\bm{v}^{T} \bm{h}_{l}\right)\left(\bm{h}_{l}^{T} \bm{v}\right) \\
     =& w_{l}\left(v_{i}-v_{j}\right)^{2}.
 \end{aligned}
 \end{equation}
which indicates that the largest connected edge can be found by maximizing $w_l(v_i-v_j)^2$, where $v_i$ and $v_j$ are the $i$th and $j$th items of Fielder vector $\bm{v}$.

Since the algebraic connectivity of a weighted graph can be measured with respect to each edge,
we can first use the graph partitioning for GNN's node classification, then if the nodes in one classification change in the later training,
we can detect them based on the greedy algorithm of algebraic connectivity, and rewire them via a link prediction method.
In real-world scenes, the dynamics and local connectivity also exhibit coupling characteristics,
therefore, one can use a coupling coefficient to measure their relationship.
Based on the above analysis on global skeleton and local link rewiring, we now present the new GNN architecture design in Fig.\ref{fig:Frameworks}.

\section{Experiments}\label{sec:experiments}
This section provides some empirical evaluations for the proposed architecture design via node classification tasks (the datasets and parameters is outlined in the Appendix).

\subsection{Model Contraction}
\begin{figure*}[htpb!]
        \centering
        \includegraphics[width=\textwidth]{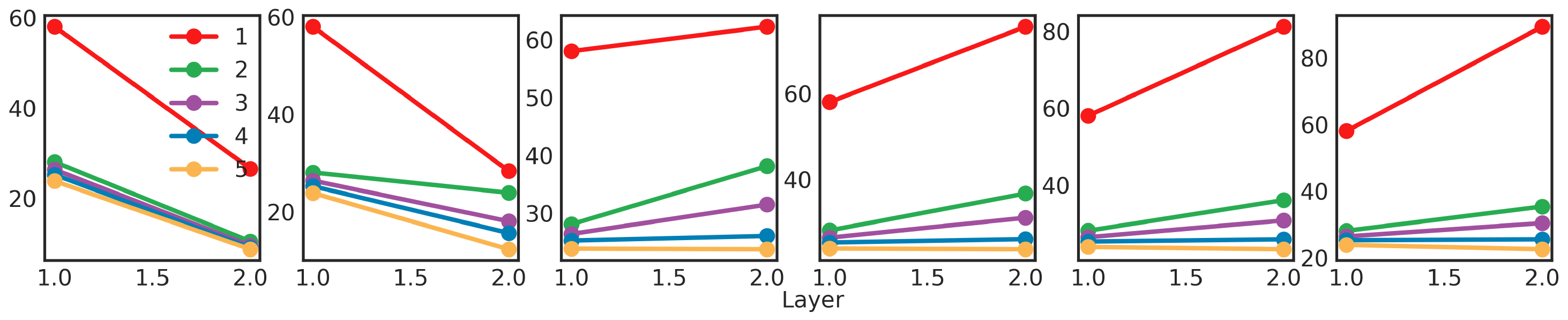}\\
        \caption{Evolutionary dynamics in training process.
        Each figure contains the $5$ prominent eigenvalues of hidden states.
        }
        \label{fig:EvolutionDynamics}
\end{figure*}

 we Fig. \ref{fig:ActivationCompare} show the model contraction properties, where the results are based on $20$ Monte Carlo experiments.
Subfigure(above) show the network width after automatic pruning.
Here we indeed observe a layer-by-layer shrinkage width, confirming our proposed shrinkage property when the depth increases. These contraction ratios,  however, seem to be relatively small.
Subfigure(bottom) compare the learning rates  on test datasets, we find that the convergence continues to decrease even adopting a large rate (i.e., 0.2, 0.5),
which illustrates that shrinkage structure can overcome the saddle point problem, and ultimately improves the convergence.
To enhance the interpretability of GNN, Fig. \ref{fig:EvolutionDynamics} demonstrates the evolutionary dynamics with respect to $5$ prominent eigenvalues of hidden states.
We observe the eigenvalues of each layer conversation from descending to ascending during the training procedure, confirming the proposed information transfer in dynamics.
\begin{figure*}
        \centering
        \begin{subfigure}[b]{0.9\textwidth}
            \centering
            \includegraphics[width=\textwidth]{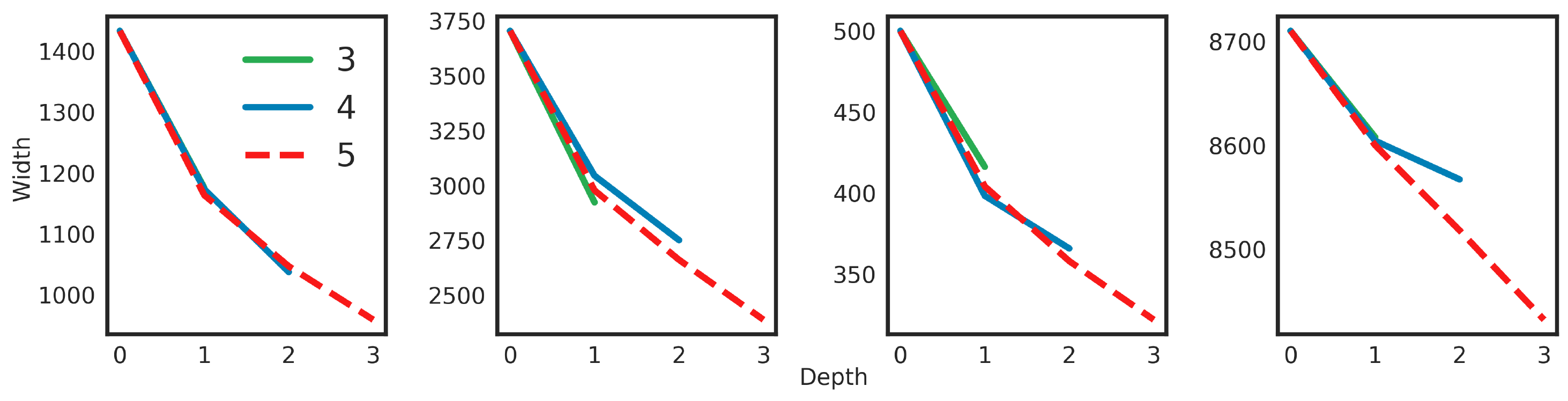}
        \end{subfigure}
        \begin{subfigure}[b]{0.9\textwidth}
            \centering
            \includegraphics[width=\textwidth]{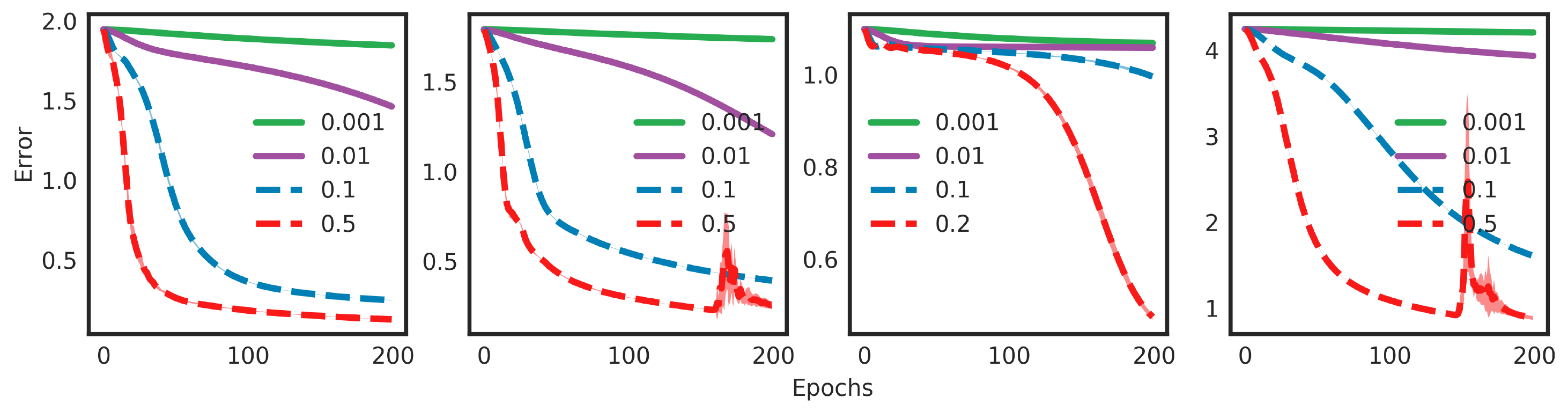}
        \end{subfigure}
        \caption[]
        {\small Performance analysis of the proposed framework.
        (Above) the observed shrinking property of network width after automatical pruning.
        (Bottom) the impact of learning rate on convergence.}
        \label{fig:ActivationCompare}
    \end{figure*}
\subsection{Multi-agent Consensus-based Link's Weight Rewiring}\label{subsec:LinkRewiring}
\begin{figure*}[htpb!]
    \centering
    \includegraphics[width=\textwidth]{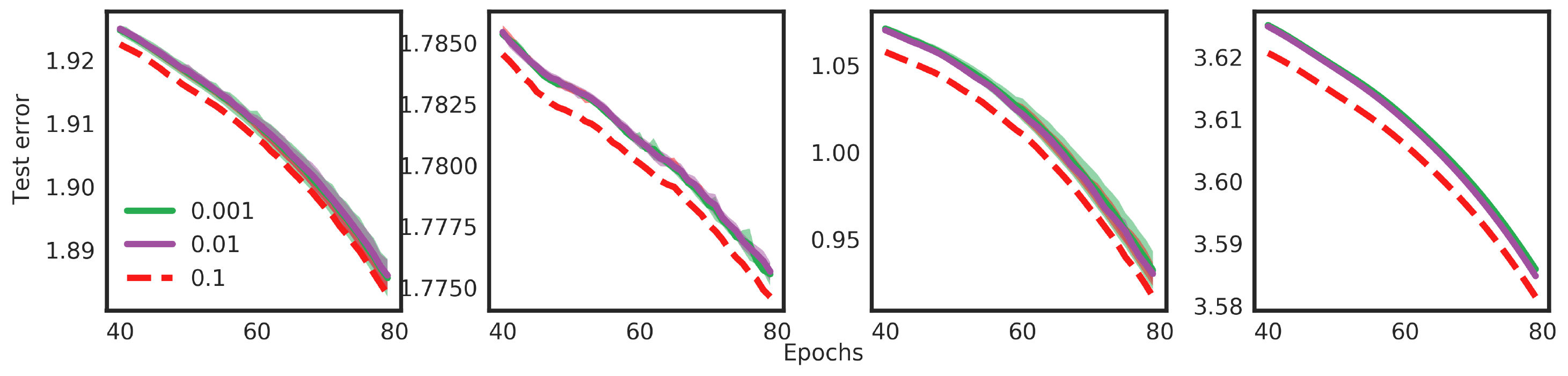}\\
    \caption{The effect of coupling coefficient on convergence.}
    \label{fig:LinkRewiring}
\end{figure*}

This section investigates the performance of multi-agents consensus-based link's weight rewiring.
Fig. \ref{fig:LinkRewiring} shows the effects of coupling coefficients on the convergence of test error.
We see the link rewiring on Citeseer, Pubmed and CoraFull are clear, while it is not obvious on Cora datasets, since the erroneous weights are not obvious.
Tab. \ref{Tab:AlgoComp} shows the test accuracy of different graph classification methods.
The results show our shrinkage-rewiring structure (SRGCN and SRChebNet) could greatly improve the node classification accuracy after automatic width pruning.

\begin{table}
\vspace{0in}
\centering
\begin{tabular}{l||c|c|c|c}
\toprule \toprule
Method&\textbf{Cora} & \textbf{Citeseer} & \textbf{Pubmed} &\textbf{CoraFull}\\
\midrule
DeepWalk &67.2 & 43.2 &65.3 &80.3\\
GAT  &56.8 &72.5 &79.0  &82.5\\
ChebNet & 62.1 &69.8 &74.4&81.4\\
GraphSAGE &64.2 &70.6 &70.5&82.2\\
Planetoid &75.7 &64.7 &77.2&80.5\\
GPNN & 68.1 &79.0 &73.6 &80.4\\
MPNN &72.0& 64.0&75.6&79.8\\
GCN &76.5  &81.5  &79.0 &86.6\\
SRGCN (ours)&80.45  &83.43  &80.16&87.13\\
SRChebNet (ours)&79.96 &82.93&80.98 &86.09\\
\hline
\end{tabular}
\centering
\caption{Performance comparison of different graph classification methods.}\label{Tab:AlgoComp}
\end{table}
\section{Related Works}\label{sec:RelatedWorks}
More recently, many innovative GNN frameworks have been developed.
Notable methods include gated GNN \cite{li2016detecting}, GraphSAGE \cite{hamilton2017inductive}, message-passing neural networks \cite{GilmerSRVD17}, and pruning networks \cite{0022KDSG17}.
In terms of architecture design on topological spaces, the most related work to ours is that Li et al. \cite{LiGDVK19}, where the authors established the equivalence between GNN and graph matching, and emphasized modeling in GNN was a convex optimization process.
The major difference is that their work does not provide a specific network skeleton design, while our method provides a shrinkage network skeleton.
Various regularization methods performed by randomly deleting hidden weights or activations are all for forming convex sets \cite{srivastava2014dropout, Rodriguez0CGR17}.
 Zhang \cite{ZhangSS19}
showed that the success of several recently proposed architectures (e.g.,  ResNet, Wide ResNet, Xception, SqueezeNet, and Inception) was mainly related to the fact that multi-branch structures help reduce the non-convex property of network topology.i
Regarding the observed modularity (clustering)  features in weight evolution,
several authors suggested defining convolutional neural network or recurrent neural network modules composed of topologically identical or similar blocks to simplify the topology design.
Results illustrated these methods could achieve a large compression ratio in terms of parameters with excellent performance guarantees \cite{zoph2018learning}.
Compared to their works, our method offers a theoretical explanation for the observed modularity phenomenon, and further employ it as an informational redundancy to guarantee local topological accuracy.
\section{Concluding Remarks}\label{sec:conclusions}
This paper presents a three-pipeline training framework based on global criticality and local topological connectivity.
From the critical Theorem on topological spaces, to reach the critical state, input and network structure should match to build a convex matching (optimization).
In specific training, to promote the information transfer under the over-parameterized setting, we propose a layer-wise shrinkage topological structure to prevent the proliferation of saddle points in high dimensional spaces.
In facing actual erroneous inputs, we give a robust topological link rewiring method based on the local connectivity required by cluster consensus, which is similar to the idea of self-supervised learning that applies structural information as redundant information for self-checking.
Our work contributes by shedding light on the success of GNNs from dynamics and topological spaces aspect.
Due to current topological structure constraints, this paper only involves the intra-layer erroneous weight rewiring, the inter-layer link imputation is still unresolved.
Further exploiting the modularity in more general topological architecture and more complex data (e.g., attacked data) is our next concern, which may provide guidelines to approach the critical expressivity.

\bibliographystyle{unsrt}
\bibliography{arxiv_Topooptimization}

\appendix
\section{Appendix}
\subsection{Schauder Fixed Point Theorem}
\begin{myDef} \label{FixedPointTheorem}\textbf{(Schauder Fixed Point Theorem)}
    Let $S$ be a local topological vector space, and  $X\subset S$ be a  closed, non-empty, bounded and convex set,
    given any continuous self-mapping $f: X \rightarrow X$, there exists a fixed point satisfing $f(x)=x$.
\end{myDef}

\subsection{Low-Rank Matrix Recovery for Rank Determination}
For the input $\bm{X}^{V\times D}$, we assume its rank $R\ll \min \{V, D\}$,  and let $\bm{A}$ be a linear map from $\mathbb{R}^{V\times D} \rightarrow \mathbb{R}^M$.
The purpose of low-rank matrix recovery is to recover $\bm{X}$ from the measurement vector $\bm{y} = \bm{A}(\bm{X})\in \mathbb{R}^{M}$.
As in the vector case, this can be achieved by solving the following problem,
\begin{equation}\label{eq:rankConstraint}
\operatorname{min} \,\operatorname {rank} ({\bm{X}}) \le R\quad {\textrm{subject to }}\bm{A}({\bm{X}}) = {\bm{y}}.
\end{equation}
The $\operatorname{rank}(\cdot)$ operator equals the $L_0$-norm of $\bm{X}$.
Computing the best low-rank approximation is analogous to the truncated singular value decomposition (SVD):
compute the SVD of the matrix, retain the larger singular values while removing the smaller ones, and then reconstruct.
The truncated SVD achieves the best approximation under Frobenius norm, which is also called the Eckart-Young theorem.
A variant of \eqref{eq:rankConstraint} is its Lagrangian form:
\begin{equation}\label{eq:low-rank_constraint}
\underset{\bm{A}}{\operatorname{min}}\left\|\bm{y}-\bm{A}\bm{X}\right\|_{F}^{2}+\alpha \cdot\operatorname{rank}(\bm{A}),
\end{equation}
where $\lVert \cdot \lVert_F$ is the Frobenius Norm and $\alpha$ is the tuning parameter. The solutions obtained for different values of $\alpha$ with $0\leq \alpha <\infty$ corresponds to the solution of (\ref{eq:rankConstraint}) obtained for $1\leq R \leq \min(V,D)$.
Unfortunately, rank minimizing for $\bm{X}$ is a non-convex and NP-hard problem due to the combinational nature of the $\operatorname{rank}(\cdot)$  operator.
Under some conditions, the solution of problem (\ref{eq:low-rank_constraint}) can be found by solving its convex relaxation:
\begin{equation}\label{eq:NuclearNorm}
\underset{\bm{A}}{\operatorname{min}}\left\|\bm{y}-\bm{A}\bm{X}\right\|_{F}^{2}+\alpha\|\bm{X}\|_{*},
\end{equation}
where $\lVert \cdot \rVert_*$ is the nuclear norm, which is equal to the sum of singular values of $\bm{X}$.
$\lVert \bm{X} \rVert_*$ is also called trace norm when it is positive semidefinite.
Unlike the $\operatorname{rank}(\cdot)$ operator in (\ref{eq:low-rank_constraint}), $\lVert \bm{X} \rVert_*$ is a convex function, and hence can be optimized via semi-definite programming and various types of other algorithms.
One method is the singular value thresholding algorithm using a hard-thresholding or soft-thresholding operator on the singular value of a specific matrix.
For medium-scale problems, computing the SVD is tractable, with a computational complexity scaling as $\mathcal{O}(R^2 \operatorname{max}(V, D))$.

\subsection{Graph Laplacian }\label{subsec:MatrixToLaplacian}
A weighted adjacency matrix  $\bm{A} = [A_{ij}]\in \mathbb{R}^{n\times n}$ of a directed graph $\mathcal{G}$ is defined such that $A_{ij}$ is the weight  $W_{j, i}$ satisfying $A_{ij} \neq 0$ if $(j, i)\in E(\mathcal{G})$,  and $A_{ij}=0$ otherwise.
The unnormalised Laplacian matrix $\bm{L}$, is then denoted as
\begin{equation}
    \bm{L} = \bm{D} - \bm{A},
\end{equation}
where the diagonal degree matrix $\bm{D} = [d_{ij}] \in \mathbb{R}^{n\times n}$ is defined as
\begin{equation*}d_{i j}=\left\{\begin{array}{ll}
\sum_{j \in \mathcal{N}_{i}} W_{i j}, & i=j \\
0, & \text {Otherwise}.
\end{array}\right.
\end{equation*}
One popular variant of graph Laplacian is its normalized form
\begin{equation}
\bm{L}_{\mathcal{N}}=\bm{D}^{-\frac{1}{2}} \bm{L} \bm{D}^{-\frac{1}{2}}=\bm{I}-\bm{D}^{-\frac{1}{2}} \bm{W D}^{-\frac{1}{2}},
\end{equation}
where $\bm{D}^{-\frac{1}{2}}=\operatorname{diag}\left(\frac{1}{\sqrt{D_{1}}}, \frac{1}{\sqrt{D_{2}}}, \ldots, \frac{1}{\sqrt{D_{n}}}\right)$.
For simplicity, we denote the normalized $\bm{L}_{\mathcal{N}}$ as $\bm{L}$ here.
From the Definition of graph Laplacian, we know that $\bm{L1} = \bm{0}$ is always hold.
That is, $0$ is an eigenvalue of $\bm{L}$ with the corresponding eigenvector $\frac{1}{\sqrt{n}}\bm{1}$.
The Laplacian matrix $ \bm {L} $ is symmetric, and its eigenvalues and eigenvectors satisfy
 \begin{equation}\label{eq:LaplacianMatrix}
\bm{L}=\bm{U}\bm{\Lambda} \bm{U}^{T}, \bm{U}^{T} \bm{U}=\bm{I}, \bm{U} \bm{U}^{T}=\bm{I},
\end{equation}
  where $\bm{\Lambda}=\operatorname{diag}\left(\lambda_{1},\lambda_{2},  \dots, \lambda_{n}\right)$ and $\bm{U}=\left(\begin{array}{cc}\frac{1}{\sqrt{n}} \bm{1} & \tilde{\bm{U}}\end{array}\right)$ with
    $\tilde{\bm{U}}^{T} \tilde{\bm{U}}=\bm{I}_{n-1}$ and $\tilde{\bm{U}}^{T} \bm{1}=\bm{0}$.

\subsection{Algebraic connectivity of directed graphs}\label{subsec:AlgebraicConnectivity}
\begin{myDef}\label{eq:AlgebraicConenctivity}
According to Mohar, no matter graph $\mathcal{G}$ is weighted or not, there is a real vector $\bm{u}\in \mathbb{R}^n$ of unit norm that can obtain the algebraic connectivity
\begin{equation}\label{eq:miniFidelerVector}
\lambda_2(\bm{L})=\min_{\bm{u} \neq 0,\bm{1}^{T}\bm{u}=0} \frac{\bm{u}^{T}\bm{Lu}}{\bm{u}^{T}\bm{u}}.
\end{equation}
\end{myDef}
For directed graphs, the algebraic connectivity can be effectively calculated by the symmetry of the Laplacian matrix
\begin{equation}\label{eq:DirectedAlgebraicConnectivity}
\lambda_2(\bm{L})=\min _{\lVert\tilde{\bm{U}}\bm{u}\rVert=\bm{1}}\bm{u}^{T}\tilde{\bm{U}}^{T} \bm{L}\tilde{\bm{U}}\bm{u}\\
=\lambda_{\min}\left(1/2\tilde{\bm{U}}^{T}\left(\bm{L}+\bm{L}^{T}\right)\tilde{\bm{U}}\right).
\end{equation}

\subsection{The Convergence of Each Subnetwork}\label{subsec:convergenceOfSubnetwork}
Suppose a network has $K$ subnetworks, each subnetwork can be viewed as a multi-agent system consisting of $n$ interacting agents, and each agent is viewed as a node of the weighted undirected graph $\mathcal{G_{\text{sub}}}$.
Each edge $\left(\mathrm{V}_{j}, \mathrm{V}_{i}\right) \in \mathcal{E}(\mathcal{G_{\text{sub}}}(t))$  or $\left(\mathrm{V}_{i}, \mathrm{V}_{j}\right) \in \mathcal{E}(\mathcal{G_{\text{sub}}}(t))$ represents an information channel between agent $V_i$ and $V_j$ at time $t$.

    Multi-agent consensus can often be modeled as information received from its value and its neighbors
    \begin{equation}
    x_{i}(t+1)=A_{ii} x_{ii}(t)+\sum_{j \in \mathcal{N}_{i}} A_{ij} x_{j}(t),
    \end{equation}
    where $\mathcal{N}_i$ is the in-neighbor set of agent $i$.
    $A=[A_{ij}]$ is the weight matrix of GNN.
    It has matrix form
    \begin{equation}\label{eq:MultiAgentConsensus}
    \bm{x}(t+1)=\bm{A}\bm{x}(t),
    \end{equation}
    Each subnetwork converges if and only if the eigenvalues of $\bm{A}$ are bounded between $-1$ and $1$.
    This can be easily obtained by reiterating  $t$ epochs
    \begin{equation}\label{eq:IterationLimit}
        \bm{x}[t]=\bm{A}^{t} \bm{x}[0].
    \end{equation}
    According the matrix decomposition of graph Laplacian  in (\ref{eq:LaplacianMatrix}), we can rewrite  (\ref{eq:IterationLimit})  as
    \begin{equation}\label{eq:MatrixDecompLimit}
        \boldsymbol{x}[t]=\bm{U}\left(\prod_{i=1}^{\infty}\left(\bm{I}-\bm{\Lambda}\right)\right) \bm{U}^{T}\boldsymbol{x}[0]
    \end{equation}
    If the subnetwork is a connected graph, it converges to
    \begin{equation}
    \lim _{t \rightarrow \infty} x[t]=\frac{1}{N} \mathbf{1} \mathbf{1}^{T} \boldsymbol{x}[0].
    \end{equation}
    This means that each node converges to the average value collected by the whole subnetwork, i.e.,
    \begin{equation}
    \lim _{t \rightarrow \infty} x_{i}(t)=\frac{1}{N} \sum_{i=1}^{N} x_{i}(0)=x^{*}.
    \end{equation}

\subsection{Datasets and Hyperparameters}
This paper employ four standard citation network datasets for node classification as benchmarks, namely, Cora, Citeseer, Pubmed, and CoraFull.
We only allow for $20$ nodes per class to be used for training, use
$100$ nodes for verification data, and $1000$ nodes for testing.
Table \ref{tab:hyperparameters} shows the hyperparameters used in the experiments.

\begin{table}[htpb!]
    \centering
    \caption{Simulation parameters}
    \begin{tabular}{l|l}
        \hline
        Parameter& Value    \\
        \hline
        Epochs      & 200 \\
        Optimizer   & SGD\\
        Activation function & Swish\\
        Monentum  & 0.9\\
        Weight decay    & 5e-4 \\
        layer number & 2-3\\
        \hline
    \end{tabular}\label{tab:hyperparameters}
\end{table}

\end{document}